\documentclass[sigconf]{acmart}
\AtBeginDocument{%
  }

\setcopyright{acmcopyright}
\copyrightyear{2025}
\acmYear{2025}
\acmDOI{XXXXXXX.XXXXXXX}
\acmConference[MM '25]{Proceedings of the 30th ACM International Conference on Multimedia}{October 27--31, 2025}{Dublin, Ireland}
\acmISBN{978-1-4503-XXXX-X/2025/06}

\begin{document}
\title{HDGlyph: A Hierarchical Disentangled Glyph-Based Framework for Long-Tail Text Rendering in Diffusion Models}


\author{
 Shuhan Zhuang\textsuperscript{1}, Mengqi Huang\textsuperscript{1}, Fengyi Fu\textsuperscript{1}, Nan Chen\textsuperscript{1}, Bohan Lei\textsuperscript{2}, Zhendong Mao\textsuperscript{1†}\\
\textsuperscript{1}University of Science and Technology of China\\
\textsuperscript{2}Zhejiang University\\
{\tt\small \{zhuangsh,huangmq,ff142536f,chen\_nan\}@mail.ustc.edu.cn, leibohan@zju.edu.cn, \{zdmao\}@ustc.edu.cn}
}

\renewcommand{\shortauthors}{Shuhan et al.}

\begin{abstract}
  Visual text rendering, which aims to accurately integrate specified textual content within generated images, is critical for various applications such as commercial design. Despite recent advances, current methods struggle with long-tail text cases, particularly when handling unseen or small-sized text. In this work, we propose a novel \textbf{H}ierarchical \textbf{D}isentangled \textbf{Glyph}-Based framework (HDGlyph) that hierarchically decouples text generation from non-text visual synthesis, enabling joint optimization of both common and long-tail text rendering. At the training stage, HDGlyph disentangles pixel-level representations via the \emph{Multi-Linguistic GlyphNet} and the \emph{Glyph-Aware Perceptual Loss}, ensuring robust rendering even for unseen characters. At inference time, HDGlyph applies \emph{Noise-Disentangled Classifier-Free Guidance} and \emph{Latent-Disentangled Two-Stage Rendering} (LD-TSR) scheme, which refines both background and small-sized text. Extensive evaluations show our model consistently outperforms others, with 5.08\% and 11.7\% accuracy gains in English and Chinese text rendering while maintaining high image quality. It also excels in long-tail scenarios with strong accuracy and visual performance.
\end{abstract}




\maketitle

\section{Introduction}
\begin{figure}[t]
    \centering
    \includegraphics[width=\linewidth]{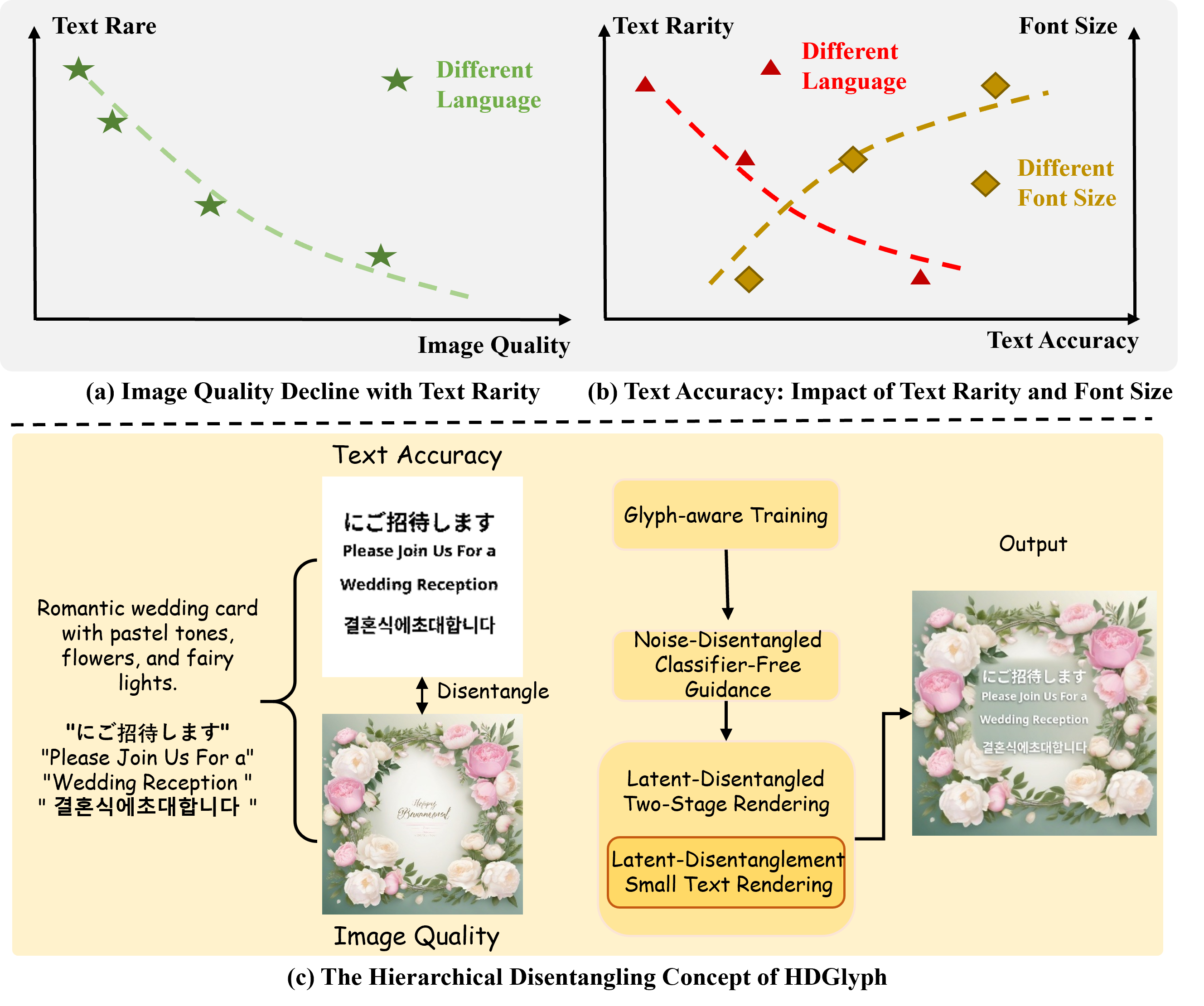}
    \caption{Illustration of our motivation. The curve chart above demonstrates the limitations of existing models on long-tail text. (a) Under the same font size, image quality decreases as the rarity of the text in the training dataset increases. (b) Under similar image quality, text accuracy decreases as the rarity of the text in the training dataset increases, and increases with the font size. For further details, refer to Appendix A. (c) Shows the hierarchical disentangling concept of HDGlyph that we propose.}
    \label{fig:introduction} 
    \Description{Provide additional visual explanation for the instruction.}
\end{figure}

Visual text rendering aims to generate images that contain accurate textual content in diverse languages and sizes, a capability that is critical for applications ranging from design materials (e.g., posters, brochures) to real-world scenarios (e.g., road signs, billboards). Although state-of-the-art models such as Stable Diffusion 3 (SD3)\cite{Esser:sd3} and FLUX\cite{flux} have demonstrated significant progress in rendering common English text, they still struggle with \emph{long-tail text rendering}, i.e., particularly non-Latin languages (e.g., Chinese, Russian, Korean) and small-sized text generation. This limitation originates from their training methodologies, that is, to ensure training stability, these models deliberately excluded training samples containing non-English languages or high-density small-sized text. Therefore, the current key in visual text rendering lies in accurately generating images capable of encompassing both long-tail text languages and text sizes.

On one hand, generating visual text that adheres to the long-tail text languages is essential for democratizing diffusion models across the world’s diverse linguistic communities. On the other hand, generating visual text that adheres to the long-tail text sizes carries tremendous commercial value, enabling the automated generation of marketing collateral, localized advertisements, and culturally tailored content at scale. Though great value, the long-tail text rendering presents significant challenges. Specifically, long-tail languages often include rare glyphs with intricate stroke structures, while rendering small text exacerbates issues such as pixelation and blurring; furthermore, the paucity of training samples for many scripts results in biased representations or outright omissions. In conclusion, overcoming these long-tail obstacles is critical for achieving truly universal visual text rendering and unlocking the full potential of diffusion models in both global and commercial contexts.

Recent works take their effort on long-tail text rendering from three aspects. The first ~\cite{Liu:Glyph-ByT5,Tuo:AnyText,tuo2024anytext2visualtextgeneration} relies on curated dataset construction to directly improve the long-tail text rendering with multilingual, in which generation performance largely depends on the dataset construction and specific model fine-tuning. Thus, other methods have been developed with greater emphasis on generalization capability. The second focus on effective training strategies, based on text-embedding optimization~\cite{ma:glyphdraw,Chen:DiffUTE,Zhao:UDiffText,Liu:Glyph-ByT5,Tuo:AnyText,Zhangli:Layout-Agnostic,tuo2024anytext2visualtextgeneration,Ma:GlyphDraw2}, or perceptual supervision in latent space~\cite{Chen:TextDiffuser,Chen:TextDiffuser-2} and image domain~\cite{Tuo:AnyText,Zhangli:Layout-Agnostic,tuo2024anytext2visualtextgeneration} to capture per-character glyph information.  The third incorporates per-character glyph conditioning and explicit layout guidance via ControlNet-style modules during generation~\cite{Yang:GlyphControl,Zhang:BrushYourText,Tuo:AnyText,Zhang:controlnet,tuo2024anytext2visualtextgeneration,Ma:GlyphDraw2}. Collectively, these approaches employ targeted training schemes and structured guidance to yield improvements in character-level accuracy.

However, we argue that existing character-learning based methods are inherently unable to disentangle the rare or small-sized glyphs from complex backgrounds, and typically lead to poor long-tail text generation performance. This stems from two key issues, on the one hand, enhancements to text encoders and the integration of OCR-based recognition accuracy into perceptual loss functions generally make models more character-aware, unseen characters at inference time frequently fall outside the learned embedding space, resulting in illegible text and degraded image quality. On the other hand, both structured-guidance approaches and the aforementioned methods capable of controlling font size suffer from pixel-level dimensionality reduction performed by convolutional layers on their input conditions, combined with the inherent structural complexity of small fonts, which undermines stroke-level clarity at reduced sizes and causes fine glyph details to blend into the background. As illustrated in Figure~\ref{fig:introduction}: (1) For text of the same size, the image quality of English text outperforms that of Chinese text, which in turn exceeds that of even rarer languages. (2) When image quality is comparable, the more long-tailed the text is, the lower its rendering quality tends to be, including cases where the font is rarer or smaller. These shortcomings underscore the need for an explicit decoupling mechanism that can simultaneously guarantee high- fidelity rendering of foreground long-tail text and faithful preservation of background context.

In this study, we propose a novel \textbf{H}ierarchical \textbf{D}isentangled \textbf{Glyph}-Based framework (HDGlyph) that hierarchically decouples text generation from non-text visual synthesis, enabling the joint optimization of both common and long-tail text rendering. During the training process, at the pixel level, HDGlyph disentangles the optimization of text and non-text content by a \emph{Multi-Linguistic GlyphNet} and a \emph{Glyph-Aware Perceptual Loss}. During the inference process, at the noise level, HDGlyph enhances glyph representation by applying \emph{Noise-Disentangled Classifier-Free Guidance} (ND-CFG) along with \emph{Latent-Disentangled Two-Stage Rendering} (LD-TSR) that preserves native image quality and further enhances glyph representation  at the latent level.

Specifically, our framework comprises three parts: 1) Glyph-aware Training Pipeline: Multi-Linguistic GlyphNet module with a Glyph-aware Perceptual Loss. This training pipeline enables a glyph-aware model that retains text rendering ability for unseen domains. Furthermore, we introduce linguistic expert LoRAs to improve language-specific performance. 2) Noise-Disentangled Classifier-Free Guidance (ND-CFG) Module: The module refines the intricate glyph details of the noise prediction process. 3) Latent-Disentangled Two-Stage Rendering (LD-TSR): During the inference process, we employ a two-stage approach. The first stage mainly ensures high-quality background rendering. The second stage mainly refines long-tail text generation and small-text rendering.

The major contributions of this work are summarized as follows:

\begin{itemize}
    \item \textbf{Concept}: For the first time, we highlight that existing paradigms fail to generate long-tail text while maintaining high-quality background synthesis due to inadequate disentanglement between text and background. We propose a novel \textbf{H}ierarchical \textbf{D}isentangled \textbf{Glyph}-Based framework (HDGlyph), which innovatively applies hierarchical disentanglement principles for visual text rendering, enabling high-quality text and background generation in challenging scenarios.
    \item \textbf{Method}: HDGlyph jointly optimizes common and long-tail text rendering by decoupling text and non-text synthesis at the pixel level via a Glyph-aware Training Pipeline, at the noise level via Noise-Disentangled Classifier-Free Guidance (ND-CFG), and at the latent level via Latent-Disentangled Two-Stage Rendering (LD-TSR).
    \item \textbf{Experiment}: Through comprehensive evaluations, our model consistently demonstrates superior performance. In common text rendering, it not only maintains high image quality, but also achieves accuracy improvements of 5.08\% in English and 11.7\% in Chinese compared to existing open-source models. Furthermore, in long-tail scenarios, such as those involving unseen characters and small text, our model exhibits particularly robust accuracy and image quality.
\end{itemize}

\section{Related Work}
\subsection{Controllable Text-to-Image Diffusion Models}
Text-to-image (T2I) diffusion models showcase impressive generative abilities, learning complex structures and meaningful semantics. However, relying solely on text prompts limits their ability to provide precise control over attributes like color and structure. Recent research addresses this by adding new conditioning methods. For example, methods such as the LoRA framework~\cite{Ye:IP-Adapter} and T2I-Adapters~\cite{Mou:T2I-Adapter} introduce lightweight modules that align internal knowledge with external control signals, enabling granular adjustments while keeping the base model intact. Similarly, IP-Adapter~\cite{Ye:IP-Adapter} enhances image-based conditioning through decoupled cross-attention, and ControlNet~\cite{Zhang:controlnet} incorporates additional encoders with zero convolution to prevent overfitting and ensure precise control. These methods have advanced applications spatial control~\cite{Mo:FreeControl}, subject control~\cite{chen2024customcontrast}, 3D generation~\cite{llein:ViewDiff} and so on.

\subsection{Visual Text Rendering}
A mainstream approach in visual text rendering focuses on supervised learning, including the enhancement of text encoders and perceptual supervision. GlyphDraw~\cite{ma:glyphdraw} fine-tunes text encoders and utilizes the CLIP image encoder for glyph embedding. DiffUTE~\cite{Chen:DiffUTE} replaces the text encoder with a pre-trained image encoder to extract glyph features. A character-level text encoder is utilized in UDiffText~\cite{Zhao:UDiffText}. Glyph-ByT5 further enhances this by fine-tuning the character-aware ByT5~\cite{Liu:Glyph-ByT5} encoder with paired text-glyph data, resulting in glyph-aligned text encoders that provide more robust text embeddings for conditional guidance. The TextDiffuser series~\cite{Chen:TextDiffuser,Chen:TextDiffuser-2} employ character-level segmentation models to supervise the accuracy of each generated character in the latent space. Another popular approach introduces glyph conditions and layout information during generation through ControlNet. GlyphControl~\cite{Yang:GlyphControl} and Brush Your Text~\cite{Zhang:BrushYourText} leverage ControlNet branches to enhance text-to-image diffusion models by utilizing glyph shape and positional information. Some methods integrate multiple strategies. For example, AnyText~\cite{Tuo:AnyText} combines auxiliary latent modules and text embedding modules in the diffusion pipeline, improving text accuracy through text-aware loss functions.  SceneTextGen~\cite{Zhangli:Layout-Agnostic} adopts character-level encoders, complemented by character-level instance segmentation models and word-level recognition models. It addresses text generation inaccuracies through context-consistency loss. AnyText2~\cite{tuo2024anytext2visualtextgeneration} introduces a WriteNet+AttnX architecture and a Text Embedding Module, enabling per-line customization of multilingual text attributes during generation and editing. GlyphDraw2~\cite{Ma:GlyphDraw2} leverages large language models (LLMs) and employs a triple-cross attention mechanism based on alignment learning. However, these approaches still face limitations in long-tail text rendering.

\section{Methodology}
\begin{figure}[t]
    \centering
    \includegraphics[width=\linewidth]{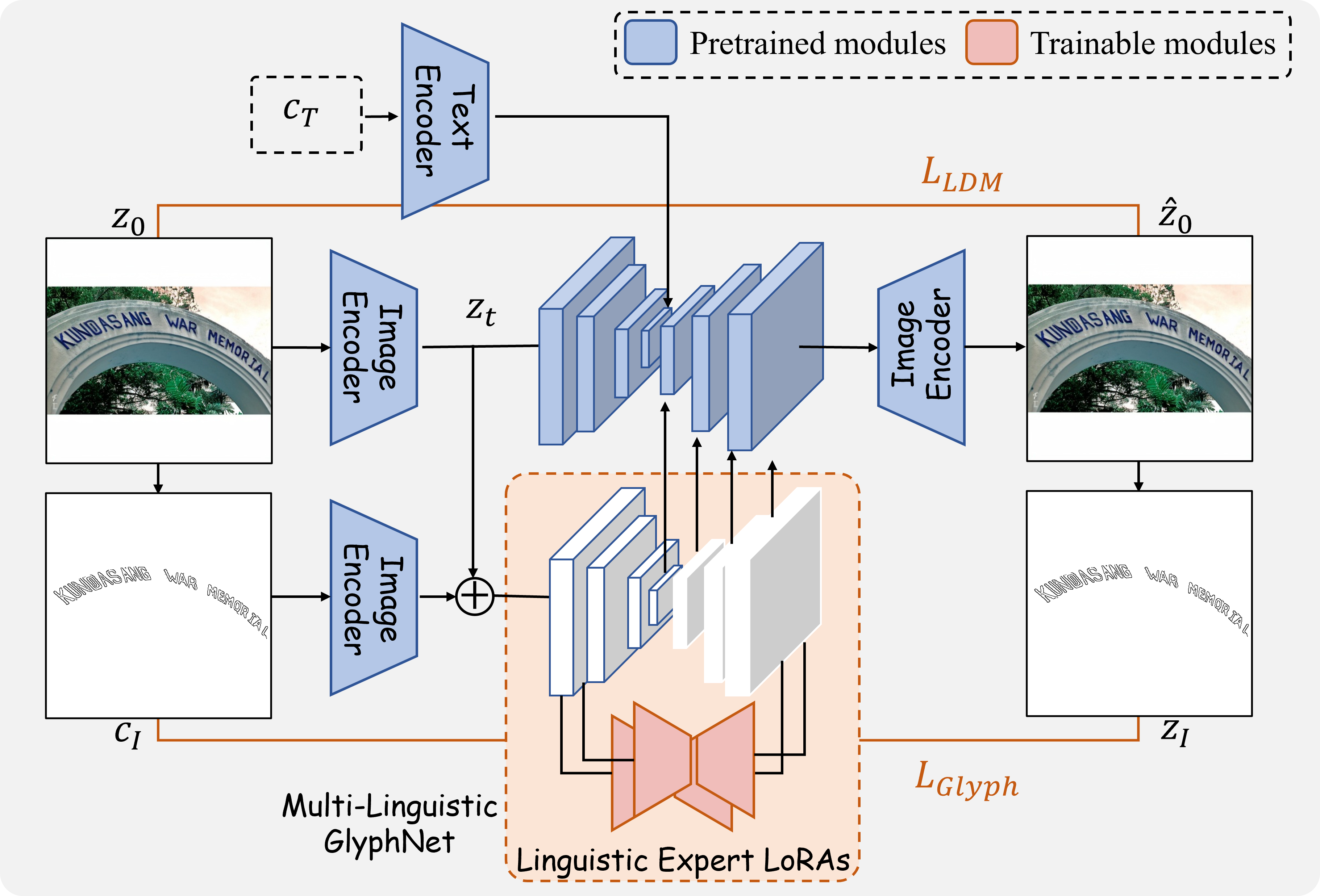}
    \caption{Glyph-aware Training Pipeline of HDGlyph.Blue modules are frozen, red modules are trainable, and white modules are styled differently to distinguish them from the U-Net.}
    \Description{Visual representation of the training pipeline of HDGlyph.}
    \label{fig:train} 
\end{figure}

\begin{figure*}
  \includegraphics[width=\textwidth]{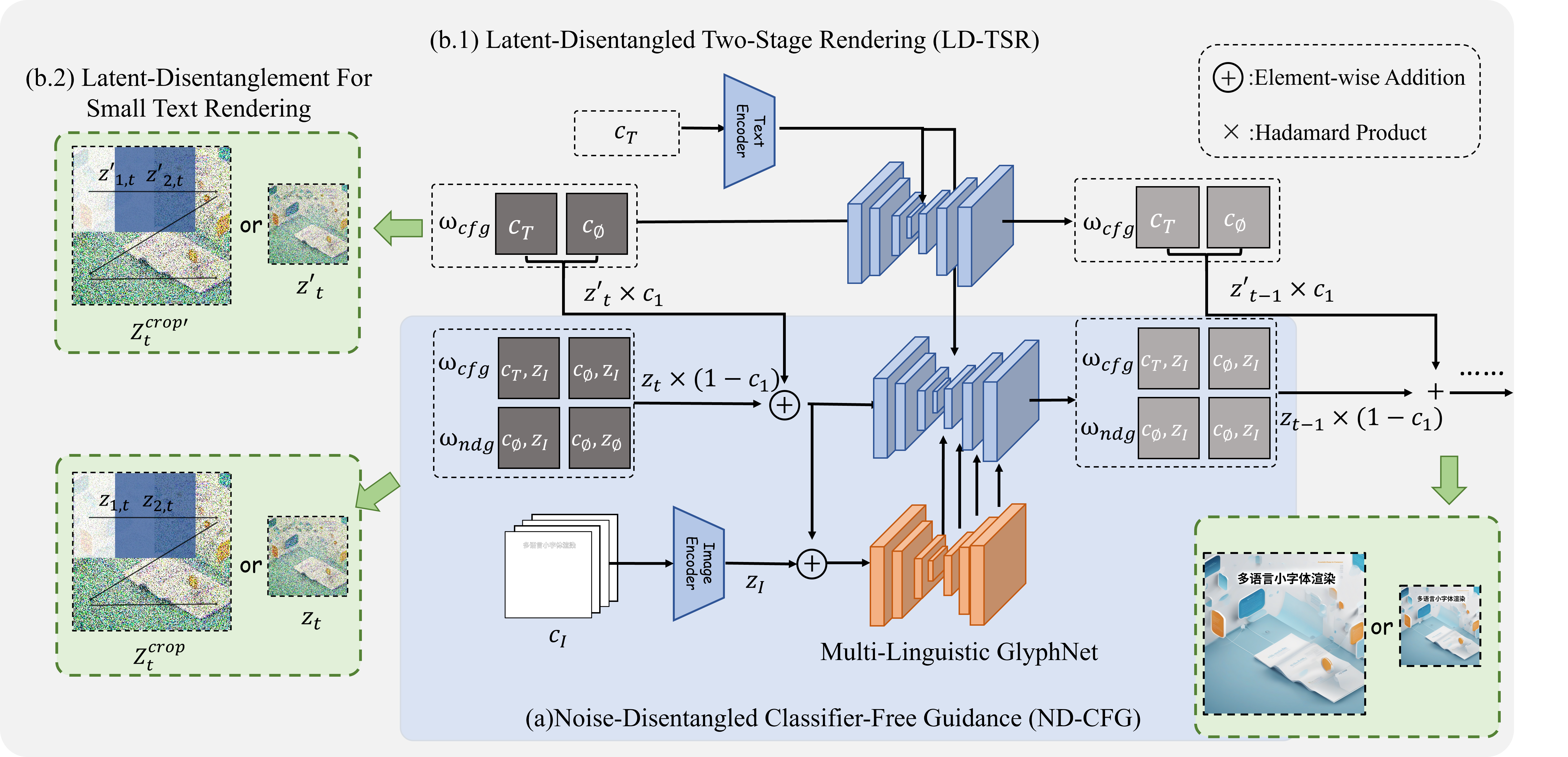}
  \caption{Our HDGlyph framework inference pipeline comprises Multi-Linguistic GlyphNet, along with (a) the Noise-Disentangled Classifier-Free Guidance (ND-CFG) module for improving glyph representation and (b.1) the Latent-Disentangled Two-Stage Rendering (LD-TSR) module for spatially separating text from the background to enhance image quality; and (b.2) the latent-disentanglement for small text rendering, which enables finer-grained glyph control at the latent level. It is noteworthy that we have omitted the process of decoding the noise image from the latent space.}
  \Description{Visual Representation of the inference pipeline of HDGlyph.}
  \label{fig:HDGlyph}
\end{figure*}
Given an input image based on glyphs $ c_I $ and a textual prompt $ c_T $, HDGlyph is designed to generate an image $ x_0 $ that faithfully encapsulates the information conveyed in $ c_I $. The overall framework of the proposed HDGlyph model is illustrated in Figure~\ref{fig:HDGlyph}. In this section, we first introduce the Glyph-aware Training Pipeline (Section 3.1), which enhances Multi-Linguistic GlyphNet's ability to leverage glyph information at the pixel level. Subsequently, we present the Noise-Disentangled Classifier-Free Guidance (ND-CFG) module (Section 3.2), which employs a classifier-guidance-inspired approach to refine the noise prediction process, thereby enhancing the glyph representation at the noise level. Finally, we discuss the Latent-Disentangled Two-Stage Rendering (LD-TSR) module (Section 3.3), which utilizes a two-stage approach to maintain high-quality background while rendering text at the latent level. Furthermore, we explore enhancing glyph representations at the latent level to improve small-sized text rendering.

\subsection{Glyph-aware Training Pipeline}  
We introduce the Multi-Linguistic GlyphNet, alongside a Glyph-aware Perceptual Loss to encourage the model to retain glyph details during generation. This pipeline ensures precise glyph generation, generalizing to unseen characters through unified multilingual structural priors and edge-aligned supervision.

\subsubsection{Multi-Linguistic GlyphNet}
Multi-Linguistic GlyphNet is designed to effectively preserve glyph structures. Part of its parameters are initialized from a pre-trained ControlNet-Canny model to leverage structural priors. As an integral component of Multi-Linguistic GlyphNet, linguistic expert LoRAs is incorporated to enhance its language-specific glyph modeling capabilities. We integrate LoRA modules at various components of GlyphControlNet, ultimately adding LoRA to the input convolutional layer and control blocks. Each language is assigned a dedicated LoRA expert, facilitating specialized feature adaptation. Ultimately, these language-specific expert LoRAs can be aggregated, allowing the model to develop a unified understanding of multiple languages and achieve comprehensive multilingual support.

\subsubsection{Glyph-Aware Perceptual Loss}
As Controlnet-Canny superior control over general object edges compared to text boundaries, we design a Glyph-Aware Perceptual Loss, which explicitly enhances the model’s sensitivity to text edges by incorporating glyph structural information. Unlike prior works \cite{Yang:GlyphControl, Tuo:AnyText, li2025joytyperobustdesignmultilingual}, which rely on OCR-based losses to enhance text recognizability, our method emphasizes consistent glyph-level alignment with the provided condition image.

During the training process of Multi-Linguistic GlyphNet, the glyph-based condition image $c_I \in \mathbb{R}^{3 \times h \times w}$ is encoded by the VAE and Multi-Linguistic GlyphNet to obtain a guidance feature $z_I$, where $h \times w$ denotes the spatial resolution in pixel space. Meanwhile, the corresponding prompt is encoded into a text embedding $c_t$ by the text encoder. A timestep $t$ is randomly selected, and noise is added to the latent representation $z_0 \in \mathbb{R}^{c \times h/8 \times w/8}$($c$ is the number of latent feature channels, and the division by 8 is due to the downsampling in the VAE.), which is obtained by encoding the original image $x_0 \in \mathbb{R}^{3 \times h \times w}$ using the VAE, resulting in a noisy latent representation $z_t$. The noise added in $z_t$ is predicted by a neural network $\epsilon_{\theta}$, conditioned on both the text embedding $c_t$ and the guidance feature $z_I$:

\begin{equation}
\label{eqn:1}
    \mathcal{L}_{\text{ldm}} = \mathbb{E}_{z_0, t, c_t, z_I, \epsilon \sim \mathcal{N}(0,1)} \left[ \|\epsilon - \epsilon_{\theta}(z_t, t, c_t, z_I)\|_2^2 \right].
\end{equation} 

Additionally, we first project the latent features back into the pixel space to reconstruct the predicted image $\hat{x}_0$ from the input image $x_0$. Leveraging the layout annotations provided in the training dataset, we extract the text regions from both the ground-truth image $x_0$ and the predicted image $\hat{x}_0$. These extracted regions are subsequently processed using a Canny edge detector to generate glyph edge maps. The Glyph-Aware Perceptual Loss $\mathcal{L}$ is formulated as follows:

\begin{equation}
\label{eqn:2}
\begin{aligned}
\mathcal{L}_{\text{Glyph}} &= \frac{\phi(t)}{hw} \sum_{h,w} \| \hat{m}_c - m_c \|^2_2, \\
\mathcal{L} &= \mathcal{L}_{\text{LDM}} + \mathcal{L}_{\text{Glyph}},
\end{aligned}
\end{equation}
where $\hat{m}_c$ and $m_c$ represent the edge maps of the predicted and ground truth glyph edge maps, respectively. The weighting function $\phi(t)$ is designed to adaptively regulate the loss contribution at different timesteps $t$, as text quality in the predicted image $\hat{x}_0$ is closely linked to the timesteps $t$. As $t$ increases, reconstructing the original image becomes more challenging, leading to greater inaccuracies in the predicted results. We apply a noise-aware schedule $\phi(t)$ to adjust the weight along with the diffusion process, such as $\phi(t) = \bar{\alpha}_t$ from DDPM \cite{ho2020denoisingdiffusionprobabilisticmodels}, or $\phi(t) = (1 - \sigma_t)^2$ as used in SD3 \cite{Esser:sd3}. Both emphasize lower-noise steps and empirically lead to improved perceptual quality.

\subsection{Noise‑Disentangled Classifier-Free Guidance} 
Classifier-free guidance enables the generation of conditional samples from an unconditional model by leveraging the unconditional score function $p(z_t)$ and a classifier $p(y | z_t)$, by sampling from the conditional distribution 
\begin{math}
    p(z_t | y) \propto p(y | z_t) p(z_t)
\end{math}~\cite{Ho:cfg}.
While sampling, we can incorporate classifier guidance by modifying $\epsilon_\theta$:
\begin{equation}
\label{eqn:3}
    \tilde{\epsilon}_\theta = \epsilon_\theta(z_t,t) + \omega_{cfg}(\epsilon_\theta(z_t,t,y) - \epsilon_\theta(z_t,t)),
\end{equation}
where $\omega_{cfg}$ controls how strongly the model adheres to the prompt by scaling the difference between the conditional and unconditional noise predictions during sampling. When a conditioning image $ c_I$ is added via ControlNet:
\begin{equation}
\label{eqn:4}
    \tilde{\epsilon}_\theta = \epsilon_\theta(z_t,t,c_I) + \omega_{cfg}(\epsilon_\theta(z_t,t,y,c_I) - \epsilon_\theta(z_t,t,c_I)),
\end{equation}

To achieve stronger glyph control, we introduce $\omega_{ndg}$, a parameter designed to disentangle the glyphs from the background at the noise level and to amplify the influence of the guidance feature $z_I$ derived from the glyph-based conditioning image $c_I$ in the generated image. To minimize the impact of non‑glyph information during generation, we also employ an image that contains no glyphs to obtain a corresponding feature representation $ z_\emptyset $, which serves as a negative image condition. Similarly, during the training phase, we probabilistically replace the image condition with $ z_\emptyset $ to enhance the model’s robustness and glyph-awareness. We further use $c_\emptyset$ to represent the negative prompt. This mechanism is integrated into the noise prediction process as follows:
\begin{equation}
\label{eqn:5}
\begin{aligned}
\tilde{\epsilon}_\theta &= \varepsilon_\theta(z_t,t,c_t,z_I) = \varepsilon_\theta(z_t,t,c_\emptyset,z_I) \\
&+ \omega_{ndg}\Big(\varepsilon_\theta(z_t,t,c_\emptyset,z_I) - \varepsilon_\theta(z_t,t,c_\emptyset,z_\emptyset)\Big) \\
&+ \omega_{cfg}\Big(\varepsilon_\theta(z_t,t,c_t,z_I) - \varepsilon_\theta(z_t,t,c_\emptyset,z_I)\Big).
\end{aligned}
\end{equation}

By combining classifier-free guidance with two disentangling weights ($\omega_{cfg}$ for background suppression and $\omega_{ndg}$ for glyph emphasis) and leveraging negative condition $z_\emptyset$, our ND-CFG module effectively separates glyph features from irrelevant background noise and selectively amplifies text-specific signals during the diffusion process, resulting in more accurate and robust text rendering.

\subsection{Latent-Disentangled Two-Stage Rendering}
In visual text rendering, in addition to achieving accurate text reproduction, it is equally crucial to ensure that overall image quality is not significantly compromised during the generation process. While our first two modules focus on reinforcing glyph preservation, they do not explicitly address background fidelity. To address this, we propose a two‑stage generation process to disentangle spatial information for enhanced rendering.

In the initial stage, noise prediction is conducted using Equation \eqref{eqn:3} without incorporating image conditions, resulting in the latent representation $ z'_0 $. Subsequently, a scheduler adds noise to $ z'_0 $, approximating the noisy latents $ z'_t $ for $ t \in (1, T) $ from the preceding denoising steps. These $ z'_t $ latents serve as background information for the current stage's latent $ z_t $. To mitigate information loss during diffusion, culminating in the latent representation at the current step:
\begin{equation}
\label{eqn:6}
    z_t \leftarrow c_1 \cdot z'_t + (1 - c_1) \cdot z_t,
\end{equation}
where 
\begin{math}
    c_1 = \left(1 + \cos \left(\frac{T-t}{T} \pi \right)/2 \right)^{\alpha_1}
\end{math}, the parameter $ \alpha_1 $ is a hyperparameter that adjusts the shape of the decay curve for $ c_1 $.

Owing to our robust disentanglement of text and background, we are able to employ a relatively large scaling factor $ \omega_{ndg} $ to exert stronger glyph control over image generation. As diffusion models tend to prioritize semantic synthesis during the early stages of denoising while focusing on texture refinement in the later stages. Building on this insight, we introduce a dynamic adjustment mechanism for the scaling factor $ \tilde{\omega}_{ndg} $, which progressively increases with the denoising timestep $ t $:

\begin{equation}
\label{eqn:7}
    \tilde{\omega}_{ndg} = \omega_{ndg} + A \cdot \left( 1 -  \left(1 + \cos \left(\frac{T-t}{T} \pi \right)/2 \right)^{\alpha_2} \right),
\end{equation}
where $ A $ determines the magnitude of the adjustment, and $ \alpha_2 $ governs the rate of increase. This progressive enhancement of $ \tilde{\omega}_{ndg} $ enables the latent space to gradually shift its focus from capturing high-level semantic structures in the early stages to refining fine-grained glyph details in the later stages, thereby ensuring a smoother transition and enhanced fidelity in the generated outputs.

The Latent-Disentangled Two-Stage Rendering module decouples text and background, enhancing the spatial representation of the background while enabling the generation of specified visual text $ c_I $, thereby ensuring visual consistency in the image.

\subsubsection{Latent-Disentanglement for Small Text Rendering}
Disentangling at the latent level can further enhance text rendering accuracy, with its benefits becoming especially evident when dealing with small-sized text. In such cases, a single Multi-Linguistic GlyphNet may struggle to provide the necessary level of fine-grained control over glyphs and to accurately generate all $ c_I $ simultaneously. Additionally, empirical observations indicate that increasing image resolution is essential to accommodate the finer structural details of the glyphs. To address this challenge, a divide-and-conquer strategy, utilizing latent-disentangled with enhanced resolution, is employed to achieve effective small-sized text rendering.

We employ a progressive upscaling strategy, beginning by initializing $ Z'_0 = \text{inter}(z'_0) $ through interpolation (e.g., bicubic), thereby approximating the high-resolution background information $ Z_t^{crop '} $ in the initial phase. Furthermore, we apply latent disentanglement to derive a patch-based denoising strategy~\cite{Bar-Tal:MultiDiffusion,Du:DemoFusion} with overlapping regions. At a given denoising step $ t $, starting with $ Z_t \in \mathbb{R}^{c \times H/8 \times W/8} $, where $ H > h $ and $ W > w $, we utilize a shifted cropping function $ \mathcal{S}_{crop}(\cdot) $ to extract a series of latent patches 
\begin{math}
    Z_t^{crop} = [z_{1,t}, \dots, z_{n,t}, \dots, z_{N,t}], \quad \mathbf{z}_{n,t} \in \mathbb{R}^{c \times h \times w}
\end{math}, along with corresponding background patches 
\begin{math}
    Z_t^{crop '} = [z'_{1,t}, \dots, z'_{n,t}, \dots, z'_{N,t}]
\end{math}, and the corresponding glyph map patches 
\begin{math}
    Z_I^{crop} = [z_{1,I}, \dots, z_{n,I}, \dots, z_{N,I}]
\end{math}:

\begin{equation}
\label{eqn:8}
    Z_t^{crop} \leftarrow c_1 \cdot Z_t^{crop '} + (1 - c_1) \cdot Z_t^{crop}.
\end{equation}

Based on this, for each patch, we predict noise as described by Equation~\eqref{eqn:5}. This disentangled latent method offers finer-grained control over the glyphs, ensuring the accuracy of small-sized text rendering while maintaining background consistency.

\section{Experiments}
\subsection{Implementation Details.}
To more comprehensively validate the generalization ability of our model, we implement it on both SDXL (U-Net)~\cite{Podell:SDXL} and SD3 (DiT)~\cite{Esser:sd3}. During training, we utilize the AnyText-3M dataset~\cite{Tuo:AnyText}, which includes both English and Chinese samples, to obtain corresponding expert LoRAs—each trained on four A100 GPUs. Following established practices~\cite{Ho:cfg,Zhang:controlnet}, we adopt prior training protocols by randomly dropping the prompt with a probability of 50\% and the image condition with a probability of 10\%. During inference, output images are generated at a resolution of 1024×1024 using a single A100 GPU. The parameter $ \omega_{ndg} $ is set to 5.0, while both $ \alpha_1 $ and $ \alpha_2 $ are set to 3.0 in Ours (U-Net) and to 4.0 in Ours (DiT), which requires a faster rate. The parameter $ A $ is fixed at 3.0. For inference with latent-level glyph enhancement, a maximum resolution of 2048×2048 is employed. For more details on the experimental setup, please refer to Appendix B.

\subsection{Benchmark and Evaluation Metrics}
\subsubsection{AnyText-Benchmark.} This benchmark~\cite{Tuo:AnyText} includes 1,000 images from LAION (primarily English) and Wukong (Chinese) to assess text detection and recognition, and it serves as our common text rendering evaluation dataset due to its focus on larger text regions with pre-provided bounding boxes; although most samples consist of single-word detection boxes, which may limit full-sentence evaluation, it provides a robust testbed for state-of-the-art comparisons in common scenarios.

\subsubsection{Multilingual-Benchmark.} Due to the scarcity of publicly available multilingual benchmarks, we constructed the Multilingual-Benchmark to evaluate our model's performance when handling long-tail data across multiple languages. Specifically, 100 prompts were randomly selected from the LAION subset of the AnyText-Benchmark, each consisting of an image caption, text, and text layout. The text was then translated into Chinese, Japanese, and Russian, resulting in a simple multilingual evaluation dataset that eliminates the influence of layout variations and font size differences. 

\subsubsection{Complex-Benchmark.} This benchmark~\cite{Ma:GlyphDraw2} comprises 200 prompts that integrate Chinese and English text. We utilize its Chinese text component, as the intricate character structures in this benchmark require more space for proper representation. Consequently, it serves to evaluate models' generation capabilities on small-sized, long-tail texts. Since no bounding boxes are provided, we employ GPT-4o to determine a predefined layout.

\subsubsection{Evaluation Metrics.} 
In this study, following~\cite{Tuo:AnyText, Ma:GlyphDraw2}, we adopted the first two evaluation metrics to assess the accuracy of visual text rendering, and the last two to evaluate the quality of the generated images: (1) Accuracy (ACC): This metric calculates the proportion of correctly generated characters in the output text relative to the total number of characters required. (2) Normalized Edit Distance (NED): A less stringent metric used to measure the similarity between two strings, reflecting the degree of alignment between the generated and target text. (3) CLIPScore:  This metric evaluates the alignment between the generated image and the corresponding image caption by computing the cosine similarity between the embeddings of the image and the caption, providing insights into semantic consistency. (4) HPSv2~\cite{wu2023humanpreferencescorev2}: This metric evaluates whether the generated images align with human preferences and serves as an indicator to assess the preference quality of the images.
 
\subsection{Comparison with State-of-The-Arts}

\subsubsection{Quantitative Results.} 
\begin{table*}[ht]
    \caption{The performance of ACC, NED, CLIPScore and HPSv2 on the AnyText-Benchmark. The best performance for each metric is highlighted in bold, and the second-best performance is indicated by underline.}
    \label{tab:AnyText-Benchmark}
    \centering
    \begin{tabular}{l|cccc|cccc}
        \toprule
        Methods & \multicolumn{4}{c|}{English} & \multicolumn{4}{c}{Chinese} \\
        \cmidrule(lr){2-5} \cmidrule(lr){6-9}
        & ACC$\uparrow$ & NED$\uparrow$ & CLIPScore$\uparrow$ & HPSv2$\uparrow$ & ACC$\uparrow$ & NED$\uparrow$ & CLIPScore$\uparrow$ & HPSv2$\uparrow$ \\
        \midrule
        SD3+Canny   & 78.99  & 86.29 & 86.70 & 20.45  & 64.08  & 80.48 & 75.10 & 18.30\\   
        FLUX+Canny  & 70.77  & 80.21 & 83.38 & 20.54 & 11.73  & 23.82 & 73.44 & 19.38\\
        Ours(DiT)   & \textbf{79.37}  & \textbf{90.72} & \textbf{88.67} & \textbf{22.68} & \textbf{75.00}  & \textbf{93.67} & \textbf{82.68} & \textbf{25.42}\\
        \midrule
        UDiffText                & 65.73    & 83.76     & 86.94     & 23.88         & -  & - & - & - \\
        TextDiffuser             & 58.10    & 78.46     & 86.50     & 24.01         & -  & -  & - & - \\
        GlyphControl$^\dagger$   & 52.62    & 75.29     & 85.48     & -             & 4.54  & 10.17  & 78.63 & - \\
        Glyph-ByT5$^\dagger$     & 73.07    & 83.53     & 48.02     & 25.11         & 72.27  & 77.99  & 40.05  & \textbf{26.01}  \\
        GlyphDraw1$^\dagger$     & 73.69    & 89.21     & 46.16     & 23.50         & 78.92 &84.76 &39.21 &25.55 \\
        GlyphDraw2$^\dagger$     &\underline{86.27}& 92.78& 47.96   & 24.51         & \underline{82.66}  & \underline{85.43} & 39.86 & \underline{25.89} \\
        Anytext                  & 74.27    & 89.40   & 89.60    & \underline{25.17}& 53.70  & 75.52  & \underline{80.93}  & 24.84  \\
        Anytext2                 & 83.58 &\underline{92.80} & \textbf{90.01}& 25.02 & 71.71  & 83.19  & 80.86  & 24.54 \\
        Ours(U-Net)               & \textbf{88.66}&\textbf{94.63}& \underline{89.65} & \textbf{25.34}&\textbf{83.41}   & \textbf{89.16}   & \textbf{80.94} & 24.90\\
        \bottomrule
    \end{tabular}
    
\end{table*}

First, we compare our model against other methods on the widely used AnyText-Benchmark, which focuses on common text rendering. Given the limited demand for small text in real-world scenarios, we do not incorporate latent-disentangled optimization for enhancing small-sized text rendering accuracy. For both English and Chinese, we employ the corresponding expert LoRA in our evaluation. Since some models are not open-sourced and certain comparative methods include additional post-processing training, we directly adopt selected experimental results from~\cite{tuo2024anytext2visualtextgeneration, Ma:GlyphDraw2}, denoted with $^\dagger$ in the table. As shown in Table~\ref{tab:AnyText-Benchmark}, our model achieves superior performance in both English and Chinese real-world scenarios, accurately rendering text while concurrently synthesizing high-quality, competitive backgrounds. Notably, there are significant numerical differences among the various CLIPScore models. Consequently, directly quoting values from other studies may result in considerable variations in reported CLIPScore values across models.

\begin{table*}[ht]
    \caption{The performance of various methods on the Multilingual-Benchmark across four languages is compared using the evaluation metrics: Acc, NED, CLIPScore and HPSv2. The best performance for each metric is highlighted in bold.}
    \label{tab:Multilingual-Benchmark}
    \centering
    \begin{tabular}{ll|ccc|cccc}
        \toprule
        Language & Metric          & SD3+Canny   & FLUX+Canny   & Ours(DiT)   & AnyText   & AnyText2   & Ours(U-Net) \\
        \midrule
        English  & Acc             & 74.99       & 66.00        &\textbf{76.07}    & 57.89    & 72.02      & \textbf{82.00} \\
                 & NED             & 83.23       & 77.27        & \textbf{87.75}   & 79.70    & 85.69     & \textbf{92.24} \\
                 & CLIPScore       & 84.77       & 82.44        & \textbf{86.67}  & 86.03    & 87.99      & \textbf{88.11}\\
                 & HPSv2           & 20.01      & 20.31         & \textbf{25.15}      & 24.80     & 24.82      & \textbf{25.34}\\
        \midrule
        Chinese  & Acc             & 39.96        & 4.44        & \textbf{47.89}        & 24.03  & 23.55  & \textbf{62.55}    \\
                 & NED             & 54.02        & 10.13       & \textbf{62.97}       & 39.08  & 33.77  & \textbf{76.18}     \\
                 & CLIPScore       & 60.67       & 59.27        & \textbf{78.87}       & 67.11  & 69.47      & \textbf{74.49}  \\
                 & HPSv2           & 13.59      & 15.11         & \textbf{23.11}       & 19.89   & 20.31      & \textbf{21.92}\\
        \midrule
        Japanese & Acc             & 41.61        & 10.14       & \textbf{42.17}         & 9.62  &14.69   & \textbf{59.70} \\
                 & NED             & 63.88        & 25.80       & \textbf{65.79}       & 27.12 &41.18  & \textbf{78.53} \\
                 & CLIPScore       & 60.70       & 58.38        & \textbf{77.70}     & 66.07 &69.90      &   \textbf{73.02}   \\
                 & HPSv2           & 13.82      & 14.95         & \textbf{22.99}      & 18.91       &18.85     &  \textbf{21.73}   \\
        \midrule
        Russian  & Acc             & 41.89        & 3.52        & \textbf{54.47}        & 3.95  &9.21   & \textbf{62.94}     \\
                 & NED             & 66.81        & 19.84       & \textbf{78.62}       & 27.55 &47.10   & \textbf{88.42}     \\
                 & CLIPScore       & 61.02       & 60.24        & \textbf{77.14}     & 65.43 & 68.49  &  \textbf{74.50}     \\
                 & HPSv2           & 13.46     & 16.30          & \textbf{23.56}      & 19.14   & 18.52& \textbf{22.47}\\
        \bottomrule
    \end{tabular}
\end{table*}

Building upon the results from the AnyText-Benchmark, we selected open-source models with demonstrated proficiency in Chinese text generation for further evaluation using the Multilingual-Benchmark. In this phase, we integrated both Chinese and English expert LoRAs into our model. Given that this evaluation focuses on long-tail text of unseen-character, we did not apply latent-disentangled text enhancements in this context. The consistent layout also mitigated the impact of font size on the results. As shown in Table~\ref{tab:Multilingual-Benchmark}, in a randomly selected subset of 100 English prompts from the AnyText-Benchmark, our model maintained excellent text accuracy and image quality across different OCR choices~\cite{EasyOCR}. For newly introduced texts in Chinese, Japanese, and Russian, within the U-Net framework, the maximum decline rates compared to English scenarios were 27.19\% in Acc and 17.4\% in NED for text accuracy, and 17.01\% in CLIPScore and 14.25\% in HPSv2 for image quality. These declines are smaller than those observed in AnyText (93.18\% in Acc, 65.97\% in NED, 23.95\% in CLIPScore, 23.75\% in HPSv2) and AnyText2 (87.21\% in Acc, 60.60\% in NED, 22.16\% in CLIPScore, 25.38\% in HPSv2). Notably, in terms of text accuracy within the DiT framework, similar trends were observed. This demonstrates that our model effectively maintains image quality and achieves satisfactory text generation when handling unseen characters in long-tail distribution data.

\begin{table}[ht]
    \caption{Performance of ACC, NED, CLIPScore and HPSv2 on the Complex-Benchmark. In the table, we denote latent-level disentanglement for small text rendering as LD. The best performance for each metric is highlighted in bold, and the second-best performance is indicated by an underline.}
    \label{tab:Complex-Benchmark}
    \centering
    \begin{tabular}{l|cccc}
        \toprule
        Methods & ACC$\uparrow$ & NED$\uparrow$ & CLIPScore$\uparrow$ & HPSv2$\uparrow$ \\
        \midrule
        SDXL+Canny      & 40.22      & 63.89      & 80.87   & 19.58  \\
        Anytext         & 10.60    & 33.38  & 81.11  & \underline{22.27}  \\
        Anytext2        & 13.04  & 35.41  & \underline{81.83}  & 21.92 \\
        Ours(U-Net)      & \underline{49.46}& \underline{71.80}& \textbf{82.98} & \textbf{22.62}\\
        Ours(U-Net)+LD & \textbf{54.48}& \textbf{74.43}& 81.37 & 20.96 \\
        \bottomrule
    \end{tabular}
\end{table}

In our final evaluation, we selected open-source models with demonstrated proficiency in Chinese text generation for assessment on the Complex-Benchmark, which is designed to evaluate a model's visual text rendering capability in long-tail distributions of small-sized text scenarios. We incorporated Chinese expert LoRA into our model, and given that this benchmark focuses on small-sized text scenarios, we applied the latent-disentanglement for small text rendering, we denote it as LD in the table. As shown in Table~\ref{tab:Complex-Benchmark}, it can be observed that the Chinese text samples here are also out-of-dataset. However, under the influence of small-sized text conditions, text accuracy further declines compared to the results in Table~\ref{tab:Multilingual-Benchmark}. Notably, our model exhibits the smallest drop in accuracy, particularly after incorporating LD, with a decrease of 8.07\%, which is lower than the drops of 13.43\% and 10.51\% observed in AnyText and AnyText2, respectively. Although the CLIPScore and HPSv2 are slightly lower, our model’s performance remains highly competitive, demonstrating robust text generation capability for long-tail small-sized text.
\begin{figure*}[ht]
    \centering
    \includegraphics[width=\textwidth]{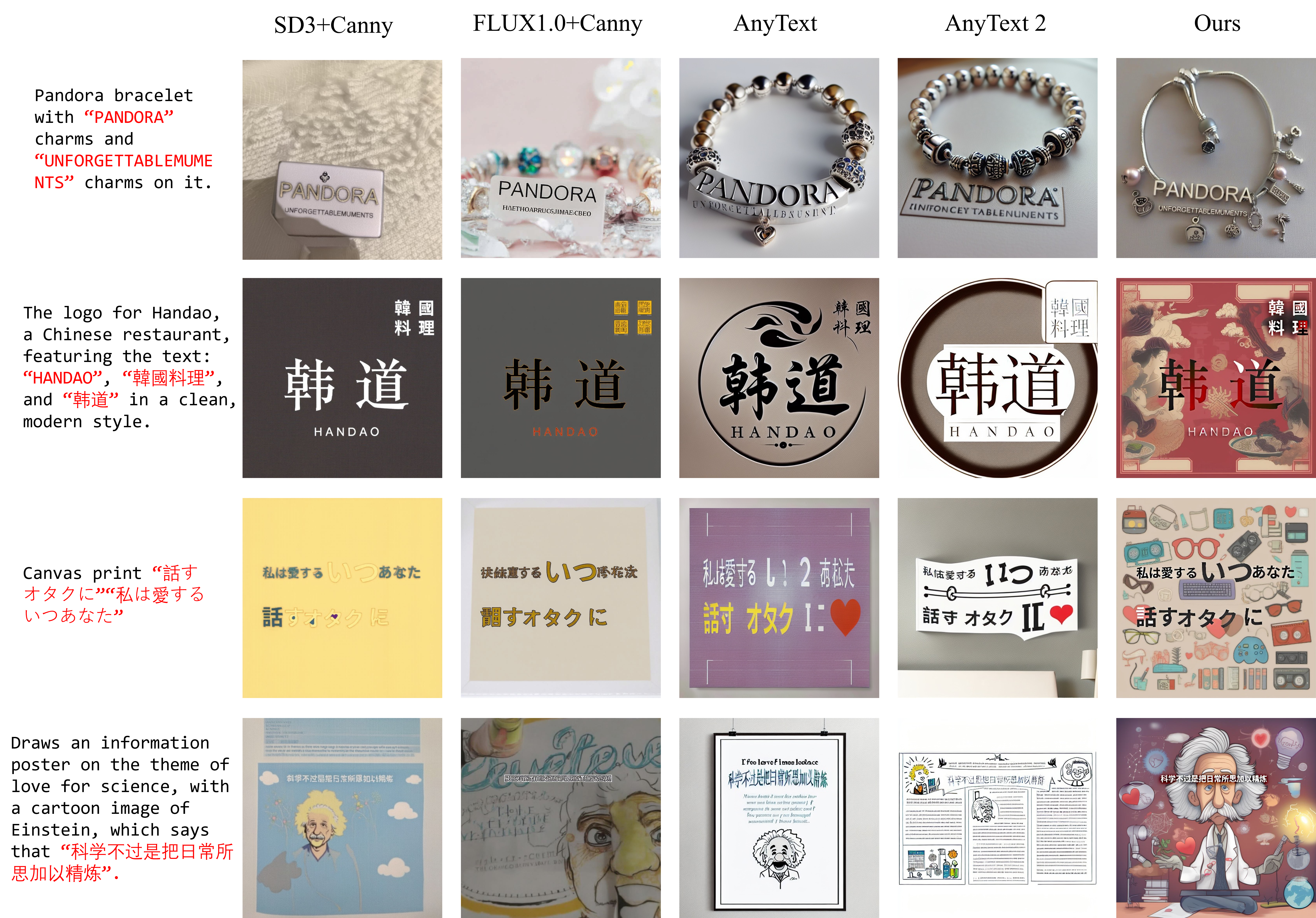}
    \caption{Qualitative comparison of HDGlyph with state-of-the-art models in long-tail text rendering of multilingual and small-sized text.}
    \label{fig:qr}
    \Description{Qualitative comparison of HDGlyph.}
\end{figure*}
\subsubsection{Qualitative Results.} As illustrated in Figure~\ref{fig:qr}, we first present the results of our model in common text rendering. To further validate the effectiveness of our model in handling long-tail text, such as multilingual content and small text commonly found in text-heavy images like posters, we also present additional relevant results. For more qualitative results, please refer to Appendix C.

\subsection{Ablation Studies}
\begin{table}[t]
    \caption{In the AnyText-Benchmark Chinese setting, we conducted ablation studies on our HDGlyph framework, evaluating its performance using Acc, NED, CLIPScore, and HPSv2. In the table, we denote latent-level disentanglement for small text rendering as LD.}
    \label{tab:ablation_HDGlyph}
    \centering
    \begin{tabular}{l|cccc}
        \toprule
        Ablation Scenario & Acc  & NED  & CLIPScore & HPSv2\\
        \midrule
        ControlNet    & 78.10    & 85.44  & 79.65 & 21.34 \\ 
        + expert LoRA      & 81.95   & 88.74   & 80.24  & 21.44    \\ 
        + ND-CFG        & \underline{84.81}   & \underline{89.25}   & 74.91 & 18.95\\
        + LD-TSR        & 83.41   & 89.16   & \textbf{80.94} & \textbf{24.90}\\
        + LD  & \textbf{85.98} & \textbf{89.80}   &\underline{80.37} & \underline{22.20}\\
        \bottomrule
    \end{tabular}
    
\end{table}
The ablation study in Table~\ref{tab:ablation_HDGlyph} demonstrates the impact of each component within the HDGlyph framework. First, the use of expert LoRA yields improvements across various metrics. When combined with the two loss functions employed during training, it slightly enhances the model’s sensitivity to glyph features and facilitates their integration into the image. This outcome is attributed to the fact that a single parameter simultaneously governs both the glyph and prompt conditions. Next, the introduction of the ND-CFG module allows for more focused enhancement of the glyph condition, resulting in an increase in accuracy by 6.71\% and NED by 3.81\% relative to the baseline, although it inevitably leads to a relative decrease in background information. The subsequent integration of the LD-TSR module overcomes this drawback. This two-stage approach not only maintains competitive text accuracy but also produces superior backgrounds and overall image quality, as evidenced by an increase in CLIPScore by 6.03\% and HPSv2 by 5.95\% compared to the model ``+ ND-CFG". Finally, we assessed the effect of the latent-level disentanglement for small text rendering. While ``+ LD" achieves higher text accuracy, the interpolation introduced during the upscaling process results in a slight decline in image quality compared to ``+ LD-TSR". This trade-off explains our decision to employ ``+ LD-TSR" in common text rendering.

\section{Conclusion}
In conclusion, our work presents the Hierarchical Disentangled Glyph-Based (HDGlyph) framework, a novel approach designed to overcome the limitations of current diffusion-based methods in visual text rendering. By integrating a multi-level disentanglement strategy with specialized modules, including the Multi-Linguistic GlyphNet, the Noise-Disentangled Classifier-Free Guidance (ND-CFG), and the Latent-Disentangled Two-Stage Rendering (LD-TSR), our framework effectively decouples the optimization of text and background generation. Comprehensive evaluations across the AnyText, Multilingual, and Complex benchmarks demonstrate that HDGlyph not only significantly improves text accuracy (with gains of 5.08\% in English and 11.7\% in Chinese) but also robustly maintains high image quality, even in challenging long-tail scenarios involving unseen characters and small-scale glyphs. These results underscore the potential of our approach to democratize visual text rendering for diverse linguistic communities and text-dense applications, setting a solid foundation for future research aimed at further expanding language coverage and refining text rendering in complex visual contexts.

\begin{acks}
To Robert, for the bagels and explaining CMYK and color spaces.
\end{acks}

\bibliographystyle{ACM-Reference-Format}
\bibliography{sample-base}


\end{document}


\title{Appendix of HDGlyph: A Hierarchical Disentangled Glyph-Based Framework for Long-Tail Text Rendering in Diffusion Models}

\maketitle
\appendix

\section{The Limitations of Existing Models on Long-Tail Text}
\subsection{The Impact of Rare Font on Visual Text Rendering}
We employ AnyText~\cite{Tuo:AnyText} and AnyText2~\cite{tuo2024anytext2visualtextgeneration} to investigate the influence of rare fonts on visual text rendering. This open-source toolkit support multilingual text generation beyond Chinese and English, making it suitable for cross-linguistic evaluation. To comprehensively assess the accuracy of visual text rendering and the perceptual quality of the generated images, we adopt three evaluation metrics: Accuracy (Acc), Normalized Edit Distance (NED), and ImageReward (RM)~\cite{xu2023imagereward}. ImageReward estimates the degree to which the generated images align with human preferences and thus serves as a proxy for evaluating perceptual quality. The inclusion of this novel human preference-based metric aims to mitigate the bias that conventional metrics may introduce into the experimental results.

\begin{table}[ht]
    \caption{The performance of various methods on the Multilingual-Benchmark across four languages is compared using the evaluation metrics: Accuracy (Acc), Normalized Edit Distance (NED) and ImageReward(RM).}
    \label{tab:Multilingual-Benchmark-2}
    \centering
    \begin{tabular}{ll|ccc}
        \toprule
        Language & Metric           & AnyText     & AnyText2  \\
        \midrule
        English  & Acc             & 57.89     & 72.02     \\
                 & NED             & 79.70     & 85.69    \\
                 & RM              & 0.8601     & 0.8585    \\
        \midrule
        Chinese  & Acc             & 24.03     & 23.55     \\
                 & NED             & 39.08     & 33.77     \\
                 & RM              & -0.8932     & -0.6697 \\
        \midrule
        Japanese & Acc             & 9.62      & 14.69    \\
                 & NED             & 27.12     & 41.18     \\
                 & RM              & -1.0556     & -0.8415\\
        \midrule
        Russian  & Acc             & 3.95     & 9.21     \\
                 & NED             & 27.55     & 47.10    \\
                 & RM              & -1.0549     &-0.9486 \\
        \bottomrule
    \end{tabular}
    
\end{table}

Since the dataset adopts a consistent layout, we assume that the font sizes are roughly the same. From the RM scores in Table~\ref{tab:Multilingual-Benchmark-2}, it can be observed that the image quality of English text outperforms that of Chinese text, which in turn exceeds that of even rarer languages such as Japanese and Russian. When the image quality differences are relatively small—such as among Chinese, Japanese, and Russian- Table~\ref{tab:Multilingual-Benchmark-2}'s Acc and NED scores reveal that the rendering quality of Chinese text is significantly higher than that of Japanese, which in turn is higher than Russian. This correlates with the rarity of these languages in the training dataset. 

\subsection{The Impact of Font Size on Visual Text Rendering}
\begin{figure*}[ht]
    \centering
    \includegraphics[width=\textwidth]{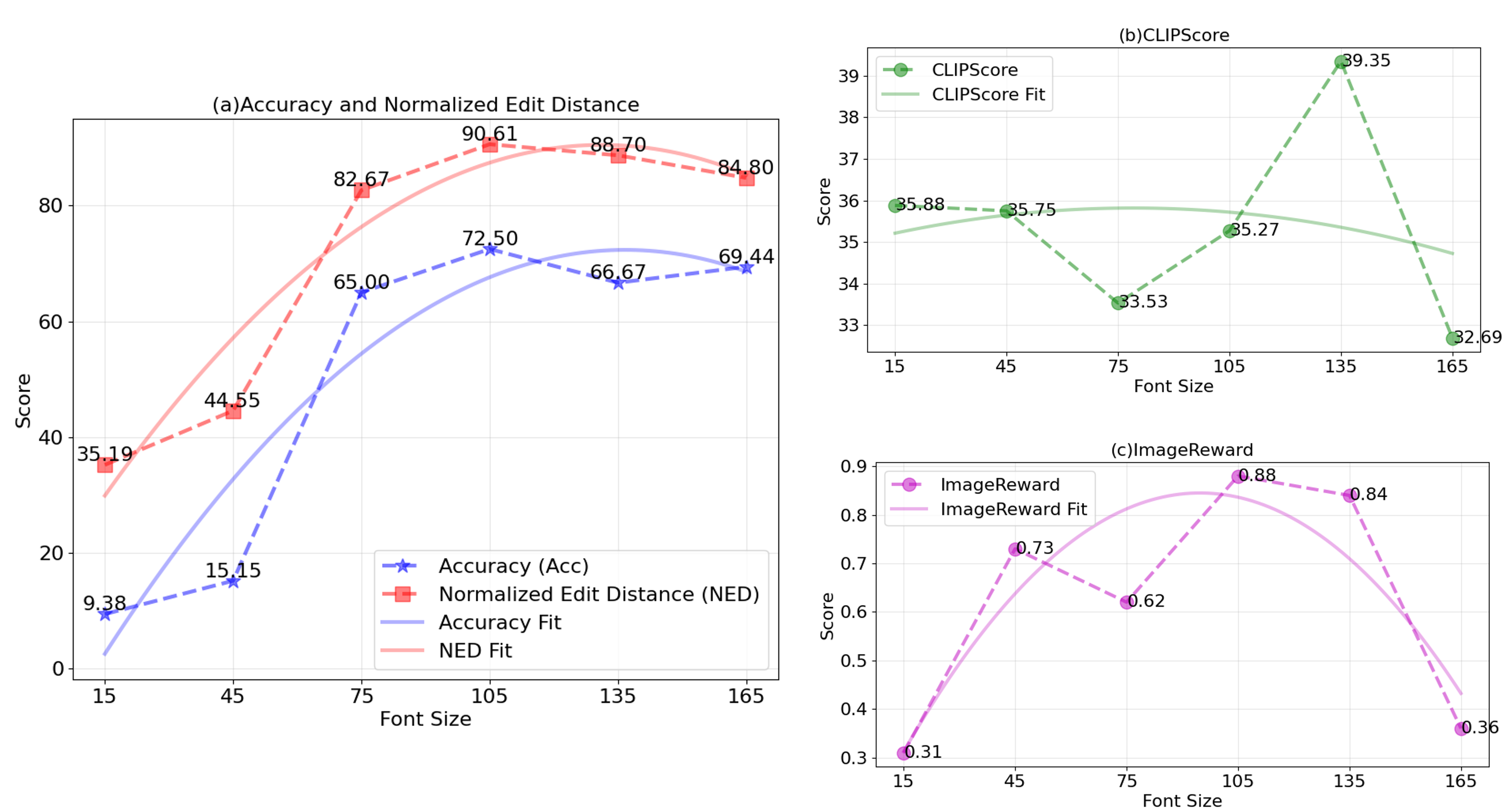}
    \caption{Using ControlNet-Canny only on the English test set of Multilingual-Benchmark (selecting the first text and its corresponding box): (a) Categorizing the data by font size and computing the corresponding Accuracy (Acc) and Normalized Edit Distance (NED) to evaluate text accuracy. (b) Categorizing the data by font size and computing the corresponding CLIPScore to evaluate image quality. (c) Categorizing the data by font size and computing the corresponding ImageReward to evaluate image quality.}
    \label{fig:limit_1}
    \Description{The Impact of Font Size on Visual Text Rendering.}
\end{figure*}

We briefly used ControlNet-Canny exclusively on the English test set of the Multilingual-Benchmark (selecting the first text and its corresponding bounding box) to demonstrate the impact of font size on visual text rendering. We evaluated text accuracy and image quality using Accuracy (Acc), Normalized Edit Distance (NED), CLIPScore (CS), and ImageReward (RM), respectively. As shown in Figure~\ref{fig:limit_1}, within the font size range of 45 to 105 where image quality remains relatively consistent (as indicated by similar CS and RM scores), we observe that as font size decreases, the quality of text rendering (i.e., Acc and NED) drops rapidly.

\section{Detail Experimental settings used in the comparison}
To ensure a fair evaluation, all methods were employed with 30 sampling steps and a CFG scale of 7.5, while the remaining parameter settings were configured according to each method's default specifications. We used the "ViT-B/32" CLIPScore model, following the setup in work~\cite{tuo2024anytext2visualtextgeneration}. To further demonstrate that the CLIPScore model yields notable differences across models, we also evaluated the results of AnyText and AnyText2 using "ViT-L/14". The CLIPScore was 0.6912 for AnyText and 0.6901 for AnyText2.

In order to eliminate the influence of different model architectures on the generated results, we categorize the models into two groups. One group comprises models with a U-Net-based architecture~\cite{Ronneberger:U-Net}, including UDiffText~\cite{Zhao:UDiffText}, TextDiffuser~\cite{Chen:TextDiffuser}, GlyphControl~\cite{Yang:GlyphControl}, Anytext~\cite{Tuo:AnyText}, Anytext2~\cite{tuo2024anytext2visualtextgeneration}, GlyphDraw~\cite{ma:glyphdraw}, GlyphDraw2~\cite{Ma:GlyphDraw2}, Glyph-ByT5~\cite{Liu:Glyph-ByT5}, and our approach (Ours (U-Net)), which is implemented on SDXL. The other group is based on DiT~\cite{William:DiT}; for this group, we mainly employ SD3 and FLUX.1, each supplemented with the corresponding Canny ControlNet for generation, to compare with our approach (Ours (DiT)) implemented on SD3. We evaluate our model using three benchmarks that progressively encompass real-world scenarios, multilingual settings, and complex small-sized text cases, with the latter two benchmarks demonstrating our model's performance on long-tail text. All values are expressed as percentages. Each prompt was used to generate four images to evaluate. 

In the AnyText-Benchmark, we followed the settings in work~\cite{Tuo:AnyText,tuo2024anytext2visualtextgeneration} to select an OCR for evaluating text accuracy. In the Multilingual-Benchmark, a multilingual OCR is required. Moreover, using different OCR systems helps eliminate the influence of OCR variations on the evaluation outcomes. Since certain works~\cite{Tuo:AnyText,tuo2024anytext2visualtextgeneration,Yang:GlyphControl} integrate PaddleOCR~\cite{PaddleOCR} as a crucial component, we employ EasyOCR~\cite{EasyOCR}, an unbiased OCR system, as the evaluation tool. This choice ensures a more impartial assessment of multilingual text accuracy, despite the potential for slightly lower numerical performance for some models.

In the first two benchmarks, we compared models from both the U-Net and DiT frameworks. However, on the Complex-Benchmark, the performance of the Latent-Disentanglement for Small Text Rendering mechanism on SD3 was suboptimal due to its reliance on features unique to SDXL—particularly its dependence on latent diffusion models’ (LDMs) inherent prior knowledge of cropped images. Since the Complex-Benchmark is primarily designed to assess long-tail small-sized text, we did not evaluate models based on the DiT framework. Future work will focus on developing more generalized approaches to enhance small-text rendering within the DiT framework.

\section{More Qualitative Results}
\begin{figure*}[ht]
    \centering
    \includegraphics[width=\textwidth]{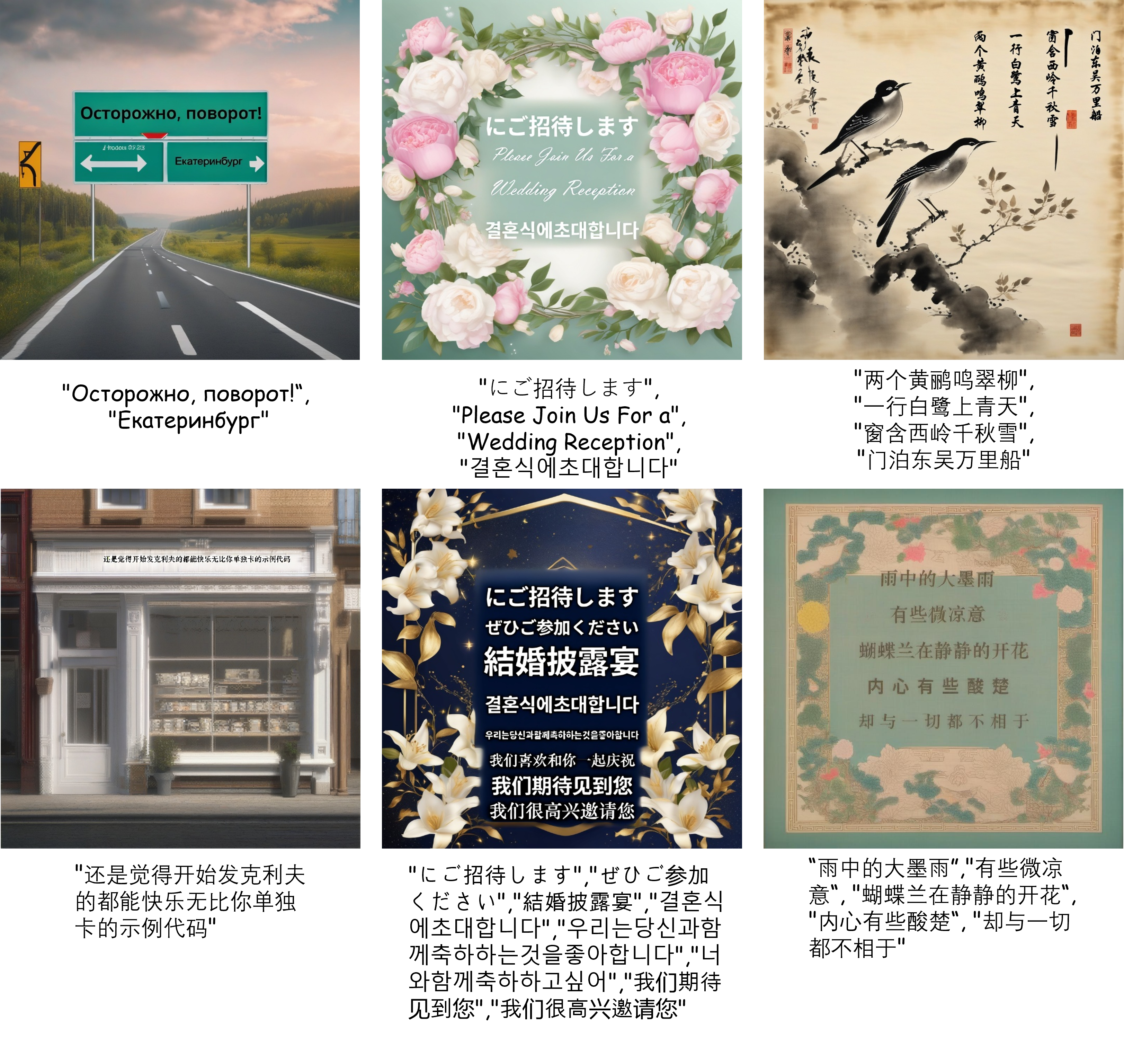}
    \caption{More Qualitative Results of HMDGlyph.}
    \label{fig:qr2}
    \Description{More Qualitative Results of HMDGlyph.}
\end{figure*}

As illustrated in Figure~\ref{fig:qr2}, we further demonstrate our model's generation performance on unseen characters and small text, underscoring its robustness in handling challenging text rendering scenarios.


\bibliographystyle{ACM-Reference-Format}
\bibliography{sample-base}








